\documentclass[runningheads]{llncs}
\usepackage[T1]{fontenc}
\usepackage{graphicx}
\usepackage{hyperref}
\usepackage{multirow}
\usepackage{graphicx}
\usepackage{nccmath}
\usepackage[table,xcdraw]{xcolor}
\usepackage[normalem]{ulem}
\usepackage{graphicx}
\usepackage[normalem]{ulem}
\useunder{\uline}{\ul}{}
\useunder{\uline}{\ul}{}
\begin{document}

\title{Optimized Custom Dataset for Efficient Detection of Underwater Trash}
%
%
\author{Jaskaran Singh Walia\inst{1}\and
Karthik Seemakurthy\inst{2}}
%

%
\institute{Vellore Institute of Technology, India \and
University of Lincoln, United Kingdom\\
\email{karanwalia2k3@gmail.com, kseemakurthy@lincoln.ac.uk}
}

%
%
\maketitle

\begin{abstract}
Accurately quantifying and removing submerged underwater waste plays a crucial role in safeguarding marine life and preserving the environment. While detecting floating and surface debris is relatively straightforward, quantifying submerged waste presents significant challenges due to factors like light refraction, absorption, suspended particles, and color distortion. This paper addresses these challenges by proposing the development of a custom dataset and an efficient detection approach for submerged marine debris. The dataset encompasses diverse underwater environments and incorporates annotations for precise labeling of debris instances. Ultimately, the primary objective of this custom dataset is to enhance the diversity of litter instances and improve their detection accuracy in deep submerged environments by leveraging state-of-the-art deep learning architectures.
The source code to replicate the results in this paper can be found at \href{https://github.com/karanwxliaa/Underwater-Trash-Detection}{GitHub.}

\end{abstract}
\section{Introduction}
\begin{figure}
    \centering
    \includegraphics[scale=0.5]{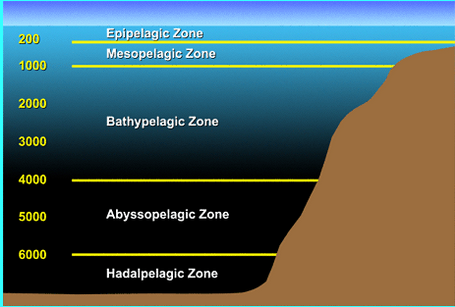}
    \caption{Zones of the Oceans~\cite{OceanLayers}.}
    \label{fig:Oceanzones}
\end{figure}

Over the past few years, the increase in underwater debris due to poor waste management practices, littering, and international industry expansion has resulted in numerous environmental issues, such as water pollution and harm to aquatic life \cite{COYLE2020100010,Derraik2002ThePO}. The debris, which remains in the epipelagic and mesopelagic zones (Fig.~\ref{fig:Oceanzones}) for years after it is dumped into the water, not only pollutes the water but also harms aquatic animals \cite{88df61e7069a431085e274d3c9068466}. However, removing debris from beneath the aquatic surface is challenging and expensive. Thus, there is a need for a cost-effective solution that can operate effectively and efficiently in a wide range of environments.

Recent advances in robotics, artificial intelligence, and automated driving \cite{10.1007/s11042-020-08976-6,https://doi.org/10.48550/arxiv.1803.10813} have made it feasible to use intelligent robots for underwater debris removal. Nevertheless, existing approaches are costly, and computationally demanding. Additionally, publicly available datasets are environment-specific, which limits their ability to produce a generalized and robust model. Therefore, we propose a new dataset where the main focus is to increase the diversity of litter instances and enhance the generalization ability of state-of-the-art object detectors. 

Autonomous underwater vehicles (AUVs) are a crucial component of a successful strategy for removing debris from maritime ecosystems. Therefore, the primary requirement for Autonomous Underwater Vehicles (AUVs) is the detection of underwater debris, specifically plastic debris. To address this challenge, we evaluated the dataset using advanced computer vision techniques to establish a baseline for litter detection. Our goal is to replace resource-intensive, time consuming algorithms with more efficient ones that will aid in real-time underwater debris detection. In this regard, we explore various deep learning based visual object detection networks that are trained and tested using the proposed dataset. The effectiveness of these detectors is evaluated using multiple metrics to validate their performance accurately.

The following are the main contributions of this paper:
\begin{itemize}
    \item Proposed a new dataset with a focus to increase the diversity of litter instances under different environments. 
    \item Trash, Rover and Bio are the three classes in the proposed dataset. 
    \item Benchmarked the litter detection by using various deep learning-based object detectors. 
\end{itemize}

\section{Related work}  {
The literature in underwater robotics has focused on the development of multi-robot systems for surface and deep water natural aquatic environment applications such as marine monitoring using learning-based recognition of underwater biological occurrences by the National Oceanic and Atmospheric Administration \cite{NOAA}. Underwater robots have also been utilized for environmental monitoring, environmental mapping \cite{temp1}, maritime robotics for guidance and localization \cite{Torrey2009Chapter1T,DREVER2018684,tftf,DERRAIK2002842}.

Underwater debris, particularly plastic waste, has become a significant environmental concern due to its detrimental effects on marine ecosystems. Plastic debris can persist in the marine environment for long periods, posing threats to marine organisms through entanglement, ingestion, and habitat destruction \cite{article111,article222}. It can also disrupt marine food webs and alter the biodiversity of marine ecosystems.

Efforts have been made to address the issue of underwater debris removal. Various methods have been employed, including manual clean-up operations, the use of remotely operated underwater vehicles (ROVs) equipped with gripping arms to physically capture debris, and the development of autonomous robotic systems specifically designed for marine debris removal. However, these approaches often face challenges in terms of efficiency, cost-effectiveness, and the vast scale of the problem. Researchers and organizations continue to explore innovative strategies and technologies to effectively tackle underwater debris and minimize its impact on the marine environment~\cite{jia2023deep,zocco2023towards}.

Recently, underwater robotics (ROVs) is considered as a popular alternative over the harmful manual methods to remove the marine debris. The vision system of a robot will aid in localising the debris and provide appropriate feedback to physically control a gripper limb to capture the objects of interest. 
A non-profit group for environmental protection and cleaning, Clear Blue Sea \cite{FRED}, has proposed the FRED (Floating Robot for Eliminating Debris). However, the FRED platform is not autonomous. In order to find garbage in marine habitats, another nonprofit, the Rozalia project, has employed underwater ROVs fitted with multibeam and side-scan sonars \cite{88df61e7069a431085e274d3c9068466}. Autonomous garbage identification and collection for terrestrial settings have also been studied, such as in the case of Kulkarni et al. \cite{kulkarni}, who employed ultrasonic devices and applied them to interior garbage. However, a vision-based system can also be envisioned.

In a study on at-sea tracking of marine detritus by Mare \cite{Girdhar2011MARE}, various tactics and technological possibilities were addressed. Following the 2011 tsunami off the shore in Japan, researchers have examined the removal of detritus from the ocean's top \cite{article1} using advanced Deep Visual Detection Models. In the study by M. Bernstein \cite{17}, LIDAR was used to locate and record beach garbage.

In recent research by Valdenegro-Toro\cite{DBLP:journals/corr/abs-2108-06800}, it was shown that a deep convolutional neural network (CNN) trained on forward-looking sonar (FLS) images could identify underwater debris with an accuracy of about 80\%. This study made use of a custom made dataset created by capturing FLS images of items frequently discovered with marine debris in a water tank. Data from water tanks were also used in the assessment.

As mentioned above, the majority of the literature which dealt with debris detection used either sonar or lidar sensors. However, visual sensors have superior resolution over sensors such as sonars or lidar. 
A sizable, labeled collection of underwater detritus is required to allow visual identification of underwater litter using a deep learning-based model. This collection needs to include information gathered from a broad variety of underwater habitats to accurately capture the various looks across wide-ranging geographic areas. There are very few datasets that are publicly available and majority of them are unlabeled. The Monterey Bay Aquarium Research Institute (MBARI) has amassed a dataset over 22 years to survey trash strewn across the ocean floor off the western coast of the United States of America \cite{4524846}, specifically plastic and metal inside and around the undersea Monterey Canyon, which traps and transports the debris in the deep oceans. The Global Oceanographic Data Center, a division of the Japan Agency for Marine-Earth Science and Technology (JAMSTEC), is another example of a publicly available large dataset. As part of the larger J-EDI (JAMSTEC E-Library of Deep-sea Images) collection, JAMSTEC has made a dataset of deep-sea detritus available online \cite{tftf}. This dataset provides type-specific debris data in the form of short video clips and images dating back to 1982. The annotated data was utilized to construct deep learning-based models for the work discussed in this paper.

In summary, various studies have been conducted on the use of autonomous robots for underwater monitoring and debris collection. The development of multi-robot systems for environmental monitoring, environmental mapping, maritime robotics, and other applications have utilized undersea robots. Researchers have also explored learning-based recognition of underwater biological occurrences for marine monitoring. Additionally, the use of remotely operated underwater vehicles (ROVs) and autonomous garbage identification and collection for terrestrial settings has been studied. A significant labeled collection of underwater trash is necessary for accurate identification using deep learning-based models.
}

\section{Dataset} \label{sec:DS}  
\subsection {\textbf{Existing datasets}}  
Although several publicly available datasets, including JAMSTEC, J-ED, TrashCAN 1.0, and Trash-ICRA19, exist for automatic waste detection, they are highly domain-specific and restricted to very limited environmental variations  \cite{MAJCHROWSKA2022274}. Table~\ref{tab:detection_existing_datsets} and \ref{tab:classification_existing_datsets} shows the statistics of existing detection and classification datasets, respectively. This limits the generalisation ability for using the vision based debris detectors across wide variety of water bodies. Also, lack of diversity in the existing datasets can induce bias into the object detection networks. The main aim of the proposed dataset is to increase the diversity of images in identifying three classes, namely Trash, Rover, and Bio, which are most useful for classifying submerged debris.

\begin{table}[h]
\centering
\resizebox{\columnwidth}{!}{%
\begin{tabular}{|l|l|l|l|l|l|}
\hline
{\color[HTML]{2E2E2E} {\ul \textbf{Type}}} &
  {\color[HTML]{2E2E2E} {\ul \textbf{Dataset}}} &
  {\color[HTML]{2E2E2E} {\ul \textbf{\# classes}}} &
  {\color[HTML]{2E2E2E} {\ul \textbf{\# images}}} &
  {\color[HTML]{2E2E2E} {\ul \textbf{Annotation type}}} &
  {\color[HTML]{2E2E2E} {\ul \textbf{Environment}}} \\ \hline
{\color[HTML]{2E2E2E} Detection} &
  {\color[HTML]{2E2E2E} Wade-AI} &
  {\color[HTML]{2E2E2E} 1} &
  {\color[HTML]{2E2E2E} 1396} &
  {\color[HTML]{2E2E2E} instance masks} &
  {\color[HTML]{2E2E2E} outdoor} \\ \hline
{\color[HTML]{2E2E2E} Detection} &
  {\color[HTML]{2E2E2E} Extended TACO} &
  {\color[HTML]{2E2E2E} 7} &
  {\color[HTML]{2E2E2E} 4562} &
  {\color[HTML]{2E2E2E} bounding box} &
  {\color[HTML]{2E2E2E} outdoor} \\ \hline
{\color[HTML]{2E2E2E} Detection} &
  {\color[HTML]{2E2E2E} UAVVaste} &
  {\color[HTML]{2E2E2E} 1} &
  {\color[HTML]{2E2E2E} 772} &
  {\color[HTML]{2E2E2E} instance masks} &
  {\color[HTML]{2E2E2E} outdoor} \\ \hline
{\color[HTML]{2E2E2E} Detection} &
  {\color[HTML]{2E2E2E} TrashCan} &
  {\color[HTML]{2E2E2E} 8} &
  {\color[HTML]{2E2E2E} 7212} &
  {\color[HTML]{2E2E2E} instance masks} &
  {\color[HTML]{2E2E2E} underwater} \\ \hline
{\color[HTML]{2E2E2E} Detection} &
  {\color[HTML]{2E2E2E} Trash-ICRA} &
  {\color[HTML]{2E2E2E} 7} &
  {\color[HTML]{2E2E2E} 7668} &
  {\color[HTML]{2E2E2E} bounding box} &
  {\color[HTML]{2E2E2E} underwater} \\ \hline
{\color[HTML]{2E2E2E} Detection} &
  {\color[HTML]{2E2E2E} Drinking waste} &
  {\color[HTML]{2E2E2E} 4} &
  {\color[HTML]{2E2E2E} 4810} &
  {\color[HTML]{2E2E2E} bounding box} &
  {\color[HTML]{2E2E2E} indoor} \\ \hline
{\color[HTML]{2E2E2E} Detection} &
  {\color[HTML]{2E2E2E} MJU-Waste} &
  {\color[HTML]{2E2E2E} 1} &
  {\color[HTML]{2E2E2E} 2475} &
  {\color[HTML]{2E2E2E} instance masks} &
  {\color[HTML]{2E2E2E} indoor} \\ \hline
{\color[HTML]{2E2E2E} Detection} &
  {\color[HTML]{2E2E2E} MJU-Waste} &
  {\color[HTML]{2E2E2E} 1} &
  {\color[HTML]{2E2E2E} 2475} &
  {\color[HTML]{2E2E2E} instance masks} &
  {\color[HTML]{2E2E2E} indoor} \\ \hline
\end{tabular}%
}
\caption{Comparison of existing litter detection datasets.}
\label{tab:detection_existing_datsets}
\end{table}
\begin{table}[ht]
\centering
\resizebox{\columnwidth}{!}{%
\begin{tabular}{|l|l|l|l|l|l|l|}
\hline
{\color[HTML]{2E2E2E} {\ul \textbf{Type}}} &
  {\color[HTML]{2E2E2E} {\ul \textbf{Dataset}}} &
  {\color[HTML]{2E2E2E} {\ul \textbf{Images}}} &
  {\color[HTML]{2E2E2E} {\ul \textbf{Classes}}} &
  {\color[HTML]{2E2E2E} {\ul \textbf{Instances}}} &
  {\color[HTML]{2E2E2E} {\ul \textbf{Environment}}} &
  {\ul \textbf{Annotation type}} \\ \hline
{\color[HTML]{2E2E2E} Classification} &
  {\color[HTML]{2E2E2E} Open Litter Map} &
  {\color[HTML]{2E2E2E} \textgreater{}100k} &
  {\color[HTML]{2E2E2E} \textgreater{}100} &
  {\color[HTML]{2E2E2E} \textgreater{}100k} &
  {\color[HTML]{2E2E2E} outdoor} &
  multilabels \\ \hline
{\color[HTML]{2E2E2E} Classification} &
  {\color[HTML]{2E2E2E} Waste pictures} &
  {\color[HTML]{2E2E2E} 23633} &
  {\color[HTML]{2E2E2E} 34} &
  {\color[HTML]{2E2E2E} 23633} &
  {\color[HTML]{2E2E2E} outdoor} &
  labels \\ \hline
{\color[HTML]{2E2E2E} Classification} &
  {\color[HTML]{2E2E2E} TrashNet} &
  {\color[HTML]{2E2E2E} 2194} &
  {\color[HTML]{2E2E2E} 5} &
  {\color[HTML]{2E2E2E} 2194} &
  {\color[HTML]{2E2E2E} indoor} &
  labels \\ \hline
\end{tabular}%
}
\caption{Comparison of existing litter classification datasets.}
\label{tab:classification_existing_datsets}
\end{table}

The Bio class provides an aspect of marine life in the environment and how much trash has affected it relative to nearby environments which can be used to further prioritise the trash cleaning. The Rover class helps differentiate the rover from being misclassified as trash in some input imagery. Finally, the Trash class helps to detect and quantify the amount of trash present in the input image/video. The dataset was curated by collecting inputs from various open-source datasets and videos across different oceans and water bodies with varying conditions and environments. We manually annotated marine debris in frames of images, focusing on selecting images with tricky object detection conditions like occlusion, noise, and illumination. We used an annotation tool \cite{7139973} to create the final dataset, which comprises of 9625 images.

\subsection{\textbf{Data Preparation}}  

\begin{figure}[!ht]
    \centering
    \begin{tabular}{ccc}
     \includegraphics[scale=0.25]{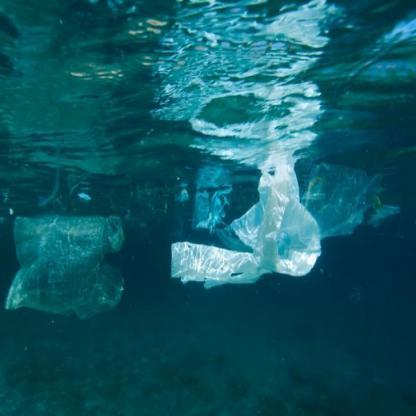} &  
     \includegraphics[scale=0.25]{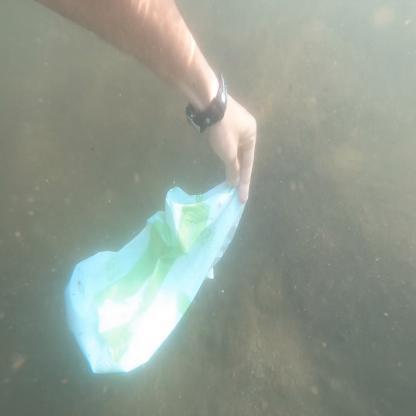} &
     \includegraphics[scale=0.25]{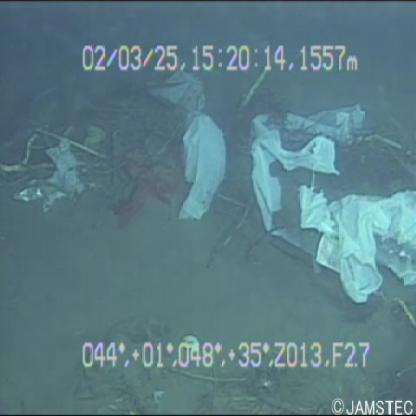} 
    \end{tabular}
    \caption{Representative images from proposed dataset.}
    \label{fig:sample_images}
\end{figure}

The first step to create this dataset is to collect inputs from various open-source datasets and videos with varying ocean environments from different countries. We manually annotated the marine debris in frames of images, focusing on selecting images with difficult object detection conditions such as occlusion, noise, and illumination. The annotations were done using a free annotation tool \cite{7139973}, resulting in 9,625 images in the dataset. A few of the sample images from our dataset are shown in Fig.~\ref{fig:sample_images}.
It can be seen that the diversity of objects and the environments that were considered in this paper. 

A Deep learning analysis was also performed on the pre-existing datasets, which can be viewed on our paper's source repository. While the models performed well on training, they failed to accurately detect classes when tested on unseen data from a slightly varying environment. Our dataset comprises of bounding box labels and image annotations and is available in more than ten different formats, making it readily importable for use with different algorithms.
The dataset was prepared using the following steps.
\begin{enumerate}

\item \textbf{Data collection}: The input images were selectively picked manually, comprising of varying environments across different regions of the world.

\item \textbf{Annotation}: The unlabelled raw images  were annotated and the annotations of labelled images were merged and renamed into three final categories, Trash, Rov and Bio which stands for underwater debris, rover (autonomous vehicle) and biological marine life respectively.

\item \textbf{Pre-processing}: These images were then rescaled to 416x416. A total of 26 classes were dropped and mapped into the final three classes. Clear water images that comprised of no annotations were also added to make the model more robust towards different environments. The dataset was further improved by randomly distorting the brightness and saturation of the images using PyTorch's built-in Transforms augmentation tool. This was done in order to mitigate the effects of spurious correlations on the model and to replicate variable underwater conditions such as illumination, occlusion, and coloring.
\end{enumerate}

The total dataset consisted of 9625 images which were split into approximately 7300 for training, 1800 for validation and 473 for test.
The Labels of the dataset were as follows:
\begin{itemize}
\item \textbf{Trash}: All sorts of marine debris (plastics, metals, etc).
\item \textbf{Bio}: All naturally occurring biological material, marine life, plants, etc.
\item \textbf{Rover}: Parts of the rover such as a robotic arm, sensors, or any part of the AUV to avoid misclassification.

The main objective behind choosing these three particular classes is that the trash class will contain all forms of trash, this increases the model's robustness when encountering unseen/new form of trash. The Bio class provides an aspect of current marine life in the environment and how much trash has affected it relative to nearby environments which can be used to prioritise the trash cleaning based on the quality of the marine life present. The Rover class helps the rover's components from being misclassified as trash in some input imagery.
\end{itemize}

\section{Benchmarking}

This section presents the latest trash detection and classification models, followed by benchmarks for the proposed dataset and statistical evaluation of the training metrics.

\subsection{\textbf{Object Detection Algorithms}}
The various architectures selected for this project were chosen from the most recent, efficient, and successful object detection networks currently in use. Each has its advantages and disadvantages, with different levels of accuracy, execution speeds, and other metrics. We utilized several state-of-the-art neural network architectures, including YOLOv7, YOLOv6s, YOLOv5s, and YOLOv4 Darknet, using their respective repositories. We also trained a custom FasterR-CNN and Mask R-CNN.

\subsection{\textbf{GPU Hardware}}
In this project, we utilized an Nvidia K80 GPU with a memory of 12 GB and a memory clock rate of 0.82GHz. This GPU was released in 2014 and has two CPU cores with 358 GB of disk space.

\subsection{\textbf{Models}} 
In this section, we discuss the latest models used and the results produced.

\subsubsection{\textbf{You Only Look Once (YOLO)}} 

You Only Look Once, or YOLO, is a popular object detection technique that can recognize multiple items in a real-time video or image. In one evaluation, it utilizes a single neural network to predict bounding boxes and class probabilities straight from the complete image. Due to this approach, YOLO is faster and more accurate than other object detection systems and therefore it can provide fast inference speeds for the real-time application of this research.

\begin{itemize}
\item \textbf{YOLOv7 tiny \cite{https://doi.org/10.48550/arxiv.2207.02696}}: The YOLOv7 algorithm outperforms its older versions in terms of speed and accuracy. It requires significantly less hardware than conventional neural networks and can be trained much more quickly on small datasets with no pre-learned weights.
\item \textbf{YOLOv5s small \cite{glenn_jocher_2022_7347926} and YOLOv6s (small) \cite{https://doi.org/10.48550/arxiv.1512.03385}}: Both of these algorithms have similar performances and results.
\end{itemize}

\subsubsection{\textbf{Faster R-CNN and Mask R-CNN}} 
Faster R-CNN and Mask R-CNN are two popular region-based object detection models that use a two-stage approach for object detection. The first stage generates region proposals, and the second stage predicts the class and refines the bounding box of each proposal.

\subsubsection{\textbf{Mask R-CNN \cite{https://doi.org/10.48550/arxiv.1512.03385}}}: Mask R-CNN extends Faster R-CNN by adding a branch to predict segmentation masks for each object. This allows the model to also segment the detected objects in addition to predicting their bounding boxes and class probabilities.

\subsection{\textbf{Evaluation metrics}} 
After the model has been trained, we utilize the testing and validation datasets, which comprise images that are mutually exclusive from the training dataset, as input to analyze the network's accuracy. The model generates a bounding box around correctly recognized items with a confidence value of.50 or higher. The amount of true positive bounding boxes drawn around marine plastic waste and true negatives serves as the basis for evaluation. 

The following performance metrics were used to validate and compare the performance of the detectors used:
\begin{itemize}
  \item \textbf{True positive and True negative values.}
  \item \textbf{Precision and Recall:} Reflects whether the model predicted debris in the input image.
  \begin{equation}
      Recall = \frac{TP}{TP + FN}, \quad Precision = \frac{TP}{TP + FP}
  \end{equation}
  
  
  \item \textbf{Mean Average Precision:} – Determines how frequently the network can correctly identify plastic. After gathering the true and false positive data, use the Intersection over Union (IoU) formula to build a precision-recall curve: 
  \begin{equation}
      IOU = \frac{BBox_{pred} \cap BBox_{GroundTruth}}{BBox_{pred} \cup BBox_{GroundTruth}}
  \end{equation}
  
Where $BBox_{Pred}$ and $BBox_{GroundTruth}$ are the expected areas under the curve for predicted and ground truth bounding boxes. To maximize accuracy, a high threshold for confidence and IoU must be specified, with a correct prediction, indicated by the threshold being exceeded. After that, the mAP can be calculated by integrating the precision-recall curve. obtained by integrating the precision-recall curve \cite{10.1007/978-3-642-40994-329}:
\begin{equation}
mAP = \int_{0}^{1} p(x) dx    
\end{equation}

\end{itemize}

 \section{Results}  

The results obtained for debris localization on our custom-curated dataset outperform previous models that used individual datasets for training. In this study, we tested the individual components of two frameworks by conducting exhaustive research on publicly available waste data in various contexts, including clean waters, natural or man-made lakes/ponds, and ocean beds. The broad range of baseline results for different contexts and diverse object dimensions will assist future researchers in this field. The tested models exhibit high average precision, mAP, and F1 scores compared to their inference speed.

The outcomes of a comprehensive study comparing several architectural networks are presented in Table~\ref{tab:results}. These trade-offs suggest that the results reported in this study better reflect long-term performance in a wider range of marine conditions, enabling a more comprehensive evaluation of the object identification model's performance in the field. Our findings suggest that YOLOv5-Small and YOLOv6s both achieve strong debris localization metrics in the real-time detection of epipelagic plastic. However, YOLOv7 yields a notably higher F1 score despite a slight reduction in inference performance. The results of a comprehensive research comparing several architectural networks is shown in the table below.
\begin{table}[ht]
\centering
\resizebox{\columnwidth}{!}{%
\begin{tabular}{|c|c|c|c|l|}
\hline
{\ul \textbf{Network}} & {\ul \textbf{mAP 0.5}} & {\ul \textbf{Precision}} & {\ul \textbf{Recall}} & {\ul \textbf{Epochs}} \\ \hline
YOLOv7       & 0.96 & 0.96 & 0.93 & 120 \\ \hline
YOLOv5s      & 0.96 & 0.95 & 0.93 & 110 \\ \hline
YOLOv6s      & 0.90 & 0.94 & 0.92 & 180 \\ \hline
Faster R-CNN & 0.81 & 0.88 & -    & 100 \\ \hline
Mask R-CNN   & 0.83 & 0.85 & -    & 100 \\ \hline
\end{tabular}%
}
\caption{Comparison between various algorithms for the purpose of benchmarking.}
\label{tab:results}
\end{table}

The trade-offs observed in our study demonstrate that the reported outcomes reflect the long-term performance of the object identification model in a wider range of marine conditions, thereby facilitating a more comprehensive evaluation of the model's performance in the field. Our findings suggest that YOLOv5-Small and YOLOv6s achieve excellent debris localization metrics in real-time detection of epipelagic plastic. However, YOLOv7 achieves a significantly higher F1 score despite a slight decrease in inference performance.

\begin{figure}
    \centering
    \begin{tabular}{cc}
     \includegraphics[height=100px, width=140px]{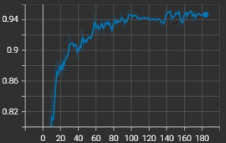} &  
     \includegraphics[height=100px, width=140px]{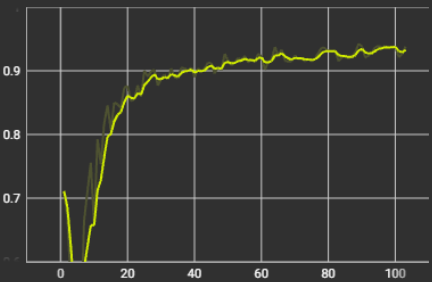} \\
     \includegraphics[height=100px, width=140px]{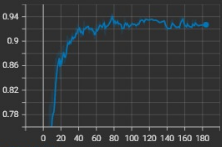} &
     \includegraphics[height=100px, width=140px]{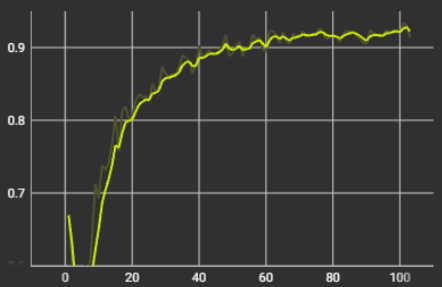} \\
       (a)  & (b)  
    \end{tabular}
    \caption{Quantitative Analysis. (a) Yolov5 and (b) Yolov8. First row: Precision curves. Second row: Recall curves.}
    \label{fig:quantitative_analysis}
\end{figure}


After evaluating multiple advanced algorithms within the same family, including YOLOv5x, v7E6E, and v8x, it was determined that the nano/small/tiny network architecture demonstrated the highest performance in evaluations, had a smaller parameter count, and required less computational power. As a result, this architecture was selected for the study. These algorithms outperformed classic Faster-RCNN and Mask-RCNN algorithms in terms of both speed and F1 score.

The performance of the model in real-world scenarios was found to be consistent with the evaluation results presented in Table 3, with only slight variations observed in a near-real-time setting. These results demonstrate the model's efficacy in identifying and categorizing underwater debris in practical applications. Furthermore, the proposed research can serve as a crucial baseline and benchmark for future investigations focused on the identification and classification of marine debris.
\begin{equation}
ImgCord_{k} = BoxScore_{i}^{j} \ast Width     
\end{equation}
where $k$ belongs to the top, down, right and left corners, $i$ is the box index, $j$ $\epsilon$  ${0, 1, 2, 3}$, and $Width$ is the image width. In the test photos, these image coordinates were utilized to illustrate the results of predicted bounding boxes.

\section{Conclusion}  
In this research, our objective was to improve object detection models by reducing dependence on environment-specific datasets. By employing our mixed, curated dataset and the latest state-of-the-art computer vision models, we were able to evaluate the feasibility of monitoring submerged marine debris in near-real-time for debris quantification. Through the use of robotic arms within Autonomous Underwater Vehicles (AUVs), our rapid inference speeds achieved a high level of performance, making real-time object detection of marine plastic litter in the ocean's epipelagic and mesopelagic layer possible, as well as the automatic detection, classification, and sorting of various submerged objects, including the collection of debris in locations such as sea-beds that are inaccessible to humans due to high pressure and other environmental factors. This application has the potential to automate trash recycling in the extreme aquatic environment with the help of deep learning. Furthermore, our proposed research serves as a fundamental baseline and benchmark for future research involving the identification and classification of underwater debris.

%
%

%
%
%
%
\bibliographystyle{elsarticle-num}

\bibliography{main}

\begin{thebibliography}{10}
\expandafter\ifx\csname url\endcsname\relax
  \def\url#1{\texttt{#1}}\fi
\expandafter\ifx\csname urlprefix\endcsname\relax\def\urlprefix{URL }\fi
\expandafter\ifx\csname href\endcsname\relax
  \def\href#1#2{#2} \def\path#1{#1}\fi

\bibitem{COYLE2020100010}
R.~Coyle, G.~Hardiman, K.~O. Driscoll,
  \href{https://www.sciencedirect.com/science/article/pii/S2666016420300086}{Microplastics
  in the marine environment: A review of their sources, distribution processes,
  uptake and exchange in ecosystems}, Case Studies in Chemical and
  Environmental Engineering 2 (2020) 100010.
\newblock \href {http://dx.doi.org/https://doi.org/10.1016/j.cscee.2020.100010}
  {\path{doi:https://doi.org/10.1016/j.cscee.2020.100010}}.
\newline\urlprefix\url{https://www.sciencedirect.com/science/article/pii/S2666016420300086}

\bibitem{Derraik2002ThePO}
J.~G.~B. Derraik, The pollution of the marine environment by plastic debris: a
  review., Marine pollution bulletin 44 9 (2002) 842--52.

\bibitem{88df61e7069a431085e274d3c9068466}
D.~Honingh, T.~{van Emmerik}, W.~Uijttewaal, H.~Kardhana, O.~Hoes, N.~{van de
  Giesen}, Urban river water level increase through plastic waste accumulation
  at a rack structure, Frontiers in Earth Science 8.
\newblock \href {http://dx.doi.org/10.3389/feart.2020.00028}
  {\path{doi:10.3389/feart.2020.00028}}.

\bibitem{OceanLayers}
Layes of the oceans,
  \url{https://www.sas.upenn.edu/msheila/biolumevolution.html}.

\bibitem{10.1007/s11042-020-08976-6}
Y.~Xiao, Z.~Tian, J.~Yu, Y.~Zhang, S.~Liu, S.~Du, X.~Lan,
  \href{https://doi.org/10.1007/s11042-020-08976-6}{A review of object
  detection based on deep learning}, Multimedia Tools Appl. 79~(33–34) (2020)
  23729–23791.
\newblock \href {http://dx.doi.org/10.1007/s11042-020-08976-6}
  {\path{doi:10.1007/s11042-020-08976-6}}.
\newline\urlprefix\url{https://doi.org/10.1007/s11042-020-08976-6}

\bibitem{https://doi.org/10.48550/arxiv.1803.10813}
J.~Andreu-Perez, F.~Deligianni, D.~Ravi, G.-Z. Yang,
  \href{https://arxiv.org/abs/1803.10813}{Artificial intelligence and robotics}
  (2018).
\newblock \href {http://dx.doi.org/10.48550/ARXIV.1803.10813}
  {\path{doi:10.48550/ARXIV.1803.10813}}.
\newline\urlprefix\url{https://arxiv.org/abs/1803.10813}

\bibitem{NOAA}
N.~Oceanic, A.~Administration,
  \href{https://oceanservice.noaa.gov/facts/marinedebris.html}{What is marine
  debris?}
\newline\urlprefix\url{https://oceanservice.noaa.gov/facts/marinedebris.html}

\bibitem{temp1}
X.~Yuan, J.-F. Mart{\'\i}nez-Ortega, J.~A.~S. Fern{\'a}ndez, M.~Eckert,
  Aekf-slam: A new algorithm for robotic underwater navigation, Sensors 17~(5)
  (2017) 1174.

\bibitem{Torrey2009Chapter1T}
L.~A. Torrey, J.~W. Shavlik, Chapter 11 transfer learning, 2009.

\bibitem{DREVER2018684}
M.~C. Drever, J.~F. Provencher, P.~D. O'Hara, L.~Wilson, V.~Bowes, C.~M.
  Bergman,
  \href{https://www.sciencedirect.com/science/article/pii/S0025326X18304259}{Are
  ocean conditions and plastic debris resulting in a ‘double whammy’ for
  marine birds?}, Marine Pollution Bulletin 133 (2018) 684--692.
\newblock \href
  {http://dx.doi.org/https://doi.org/10.1016/j.marpolbul.2018.06.028}
  {\path{doi:https://doi.org/10.1016/j.marpolbul.2018.06.028}}.
\newline\urlprefix\url{https://www.sciencedirect.com/science/article/pii/S0025326X18304259}

\bibitem{tftf}
Tensorflow, \href{https://github.com/
  tensorflow/models/blob/master/research/object detection/g3doc/ detection
  model zoo.md}{"tensorflow object detection zoo}.
\newline\urlprefix\url{https://github.com/
  tensorflow/models/blob/master/research/object detection/g3doc/ detection
  model zoo.md}

\bibitem{DERRAIK2002842}
J.~G. Derraik,
  \href{https://www.sciencedirect.com/science/article/pii/S0025326X02002205}{The
  pollution of the marine environment by plastic debris: a review}, Marine
  Pollution Bulletin 44~(9) (2002) 842--852.
\newblock \href
  {http://dx.doi.org/https://doi.org/10.1016/S0025-326X(02)00220-5}
  {\path{doi:https://doi.org/10.1016/S0025-326X(02)00220-5}}.
\newline\urlprefix\url{https://www.sciencedirect.com/science/article/pii/S0025326X02002205}

\bibitem{article111}
R.~Thompson, Y.~Olsen, R.~Mitchell, A.~Davis, S.~Rowland, A.~John,
  D.~Mcgonigle, A.~Russell, Lost at sea: Where is all the plastic?, Science
  (New York, N.Y.) 304 (2004) 838.
\newblock \href {http://dx.doi.org/10.1126/science.1094559}
  {\path{doi:10.1126/science.1094559}}.

\bibitem{article222}
J.~Jambeck, R.~Geyer, C.~Wilcox, T.~Siegler, M.~Perryman, A.~Andrady,
  R.~Narayan, K.~Law, Marine pollution. plastic waste inputs from land into the
  ocean, Science (New York, N.Y.) 347 (2015) 768--771.
\newblock \href {http://dx.doi.org/10.1126/science.1260352}
  {\path{doi:10.1126/science.1260352}}.

\bibitem{jia2023deep}
T.~Jia, Z.~Kapelan, R.~de~Vries, P.~Vriend, E.~C. Peereboom, I.~Okkerman,
  R.~Taormina, Deep learning for detecting macroplastic litter in water bodies:
  A review, Water Research (2023) 119632.

\bibitem{zocco2023towards}
F.~Zocco, T.-C. Lin, C.-I. Huang, H.-C. Wang, M.~O. Khyam, M.~Van, Towards more
  efficient efficientdets and real-time marine debris detection, IEEE Robotics
  and Automation Letters 8~(4) (2023) 2134--2141.

\bibitem{FRED}
Fred cars, \url{https://www.sandiego.edu/news/detail.php?_focus=72984}.

\bibitem{kulkarni}
S.~Kulkarni, S.~Junghare, Robot based indoor autonomous trash detection
  algorithm using ultrasonic sensors, 2013, pp. 1--5.
\newblock \href {http://dx.doi.org/10.1109/CARE.2013.6733698}
  {\path{doi:10.1109/CARE.2013.6733698}}.

\bibitem{Girdhar2011MARE}
Y.~Girdhar, A.~Xu, B.~B. Dey, M.~Meghjani, F.~Shkurti, I.~Rekleitis, G.~Dudek,
  {MARE: Marine Autonomous Robotic Explorer}, in: Proceedings of the 2011
  IEEE/RSJ International Conference on Intelligent Robots and Systems (IROS
  '11), San Francisco, USA, 2011, pp. 5048--5053.

\bibitem{article1}
M.~Fulton, J.~Hong, M.~Islam, J.~Sattar, Robotic detection of marine litter
  using deep visual detection models.

\bibitem{17}
M.~Bernstein, R.~Graham, D.~Cline, J.~M. Dolan, K.~Rajan, Learning-based event
  response for marine robotics (2013) 3362–3367.

\bibitem{DBLP:journals/corr/abs-2108-06800}
D.~Singh, M.~Valdenegro{-}Toro, \href{https://arxiv.org/abs/2108.06800}{The
  marine debris dataset for forward-looking sonar semantic segmentation}, CoRR
  abs/2108.06800.
\newblock \href {http://arxiv.org/abs/2108.06800} {\path{arXiv:2108.06800}}.
\newline\urlprefix\url{https://arxiv.org/abs/2108.06800}

\bibitem{4524846}
L.~Stutters, H.~Liu, C.~Tiltman, D.~J. Brown, Navigation technologies for
  autonomous underwater vehicles, IEEE Transactions on Systems, Man, and
  Cybernetics, Part C (Applications and Reviews) 38~(4) (2008) 581--589.
\newblock \href {http://dx.doi.org/10.1109/TSMCC.2008.919147}
  {\path{doi:10.1109/TSMCC.2008.919147}}.

\bibitem{MAJCHROWSKA2022274}
S.~Majchrowska, A.~Mikołajczyk, M.~Ferlin, Z.~Klawikowska, M.~A. Plantykow,
  A.~Kwasigroch, K.~Majek,
  \href{https://www.sciencedirect.com/science/article/pii/S0956053X21006474}{Deep
  learning-based waste detection in natural and urban environments}, Waste
  Management 138 (2022) 274--284.
\newblock \href
  {http://dx.doi.org/https://doi.org/10.1016/j.wasman.2021.12.001}
  {\path{doi:https://doi.org/10.1016/j.wasman.2021.12.001}}.
\newline\urlprefix\url{https://www.sciencedirect.com/science/article/pii/S0956053X21006474}

\bibitem{7139973}
S.~Alexandrova, Z.~Tatlock, M.~Cakmak, Roboflow: A flow-based visual
  programming language for mobile manipulation tasks, in: 2015 IEEE
  International Conference on Robotics and Automation (ICRA), 2015, pp.
  5537--5544.
\newblock \href {http://dx.doi.org/10.1109/ICRA.2015.7139973}
  {\path{doi:10.1109/ICRA.2015.7139973}}.

\bibitem{https://doi.org/10.48550/arxiv.2207.02696}
C.-Y. Wang, A.~Bochkovskiy, H.-Y.~M. Liao,
  \href{https://arxiv.org/abs/2207.02696}{Yolov7: Trainable bag-of-freebies
  sets new state-of-the-art for real-time object detectors} (2022).
\newblock \href {http://dx.doi.org/10.48550/ARXIV.2207.02696}
  {\path{doi:10.48550/ARXIV.2207.02696}}.
\newline\urlprefix\url{https://arxiv.org/abs/2207.02696}

\bibitem{glenn_jocher_2022_7347926}
G.~Jocher, A.~Chaurasia, A.~Stoken, J.~Borovec, NanoCode012, Y.~Kwon,
  K.~Michael, TaoXie, J.~Fang, imyhxy, Lorna, Z.~Yifu, C.~Wong, A.~V,
  D.~Montes, Z.~Wang, C.~Fati, J.~Nadar, Laughing, UnglvKitDe, V.~Sonck,
  tkianai, yxNONG, P.~Skalski, A.~Hogan, D.~Nair, M.~Strobel, M.~Jain,
  \href{https://doi.org/10.5281/zenodo.7347926}{{ultralytics/yolov5: v7.0 -
  YOLOv5 SOTA Realtime Instance Segmentation}} (Nov. 2022).
\newblock \href {http://dx.doi.org/10.5281/zenodo.7347926}
  {\path{doi:10.5281/zenodo.7347926}}.
\newline\urlprefix\url{https://doi.org/10.5281/zenodo.7347926}

\bibitem{https://doi.org/10.48550/arxiv.1512.03385}
K.~He, X.~Zhang, S.~Ren, J.~Sun, \href{https://arxiv.org/abs/1512.03385}{Deep
  residual learning for image recognition} (2015).
\newblock \href {http://dx.doi.org/10.48550/ARXIV.1512.03385}
  {\path{doi:10.48550/ARXIV.1512.03385}}.
\newline\urlprefix\url{https://arxiv.org/abs/1512.03385}

\bibitem{10.1007/978-3-642-40994-329}
K.~Boyd, K.~H. Eng, C.~D. Page,
  \href{https://doi.org/10.1007/978-3-642-40994-329}{Area under the
  precision-recall curve: Point estimates and confidence intervals}, in:
  Proceedings of the 2013th European Conference on Machine Learning and
  Knowledge Discovery in Databases - Volume Part III, ECMLPKDD'13,
  Springer-Verlag, Berlin, Heidelberg, 2013, p. 451–466.
\newblock \href {http://dx.doi.org/10.1007/978-3-642-40994-329}
  {\path{doi:10.1007/978-3-642-40994-329}}.
\newline\urlprefix\url{https://doi.org/10.1007/978-3-642-40994-329}

\end{thebibliography}

\end{document}